\documentclass[10pt,twocolumn,letterpaper]{article}

\usepackage{iccv}
\usepackage{times}
\usepackage{epsfig}
\usepackage{graphicx}
\usepackage{amsmath}
\usepackage{amssymb}
\usepackage[accsupp]{axessibility}  

\usepackage{bm}
\usepackage{booktabs}
\usepackage{multirow}
\usepackage[pagebackref=true,breaklinks=true,letterpaper=true,colorlinks,bookmarks=false]{hyperref}

\iccvfinalcopy 

\def\iccvPaperID{5698} 

\ificcvfinal\fi

\begin{document}

\title{MonoNeRD: NeRF-like Representations for Monocular 3D Object Detection}

\author{Junkai Xu\textsuperscript{\rm 1,2\thanks{Work performed during an internship at FABU Inc.}}\quad Liang Peng\textsuperscript{\rm 1,2,*}\quad Haoran Cheng\textsuperscript{\rm 1,2,*} \quad Hao Li\textsuperscript{\rm 2} \\ Wei Qian\textsuperscript{\rm 2} \quad Ke Li\textsuperscript{\rm 4} \quad Wenxiao Wang\textsuperscript{\rm 3\thanks{Corresponding author}} \quad Deng Cai\textsuperscript{\rm 1,2} \\
$^1$State Key Lab of CAD \& CG, Zhejiang University \quad $^2$FABU Inc.\\
$^3$ School of Software Technology, Zhejiang University \quad $^4$Fullong Inc.\\
{\tt\small \{xujunkai, pengliang, haorancheng\}@zju.edu.cn} \\
}
\maketitle
\ificcvfinal\thispagestyle{empty}\fi

\begin{abstract}
    In the field of monocular 3D detection, it is common practice to utilize scene geometric clues to enhance the detector's performance.
    However, many existing works adopt these clues explicitly such as estimating a depth map and back-projecting it into 3D space.
    This explicit methodology induces sparsity in 3D representations due to the increased dimensionality from 2D to 3D, and leads to substantial information loss, especially for distant and occluded objects.
    To alleviate this issue, we propose \textbf{MonoNeRD}, a novel detection framework that can infer dense 3D geometry and occupancy.
    Specifically, we model scenes with Signed Distance Functions (SDF), facilitating the production of dense 3D representations.
    We treat these representations as Neural Radiance Fields (NeRF) and then employ volume rendering to recover RGB images and depth maps.
    To the best of our knowledge, this work is the first to introduce volume rendering for M3D, and demonstrates the potential of implicit reconstruction for image-based 3D perception. 
    Extensive experiments conducted on the KITTI-3D benchmark and Waymo Open Dataset demonstrate the effectiveness of \textbf{MonoNeRD}.
    Codes are available at \url{https://github.com/cskkxjk/MonoNeRD}.
\end{abstract}
\section{Introduction}
\label{sec:intro}
    Monocular 3D detection (M3D) is an active research topic in the computer vision community due to its convenience, low cost and wide range of applications, including autonomous driving, robotic navigation and more. 
    The key point of the task is to establish reasonable correspondences between 2D images and 3D space. 
    Some works leverage geometrical priors to extract 3D information, such as object poses via 2D-3D constraints. 
    These constraints usually require additional keypoint annotations \cite{chen2016monocular,he2019mono3d++} or CAD models \cite{chabot2017deep, manhardt2019roi}. 
    Other works convert estimated depth maps into 3D point cloud representations (Pseudo-LiDAR) \cite{ma2020rethinking,wang2019pseudo,you2019pseudo}. 
    Depth estimates are also used to combine with image features \cite{xu2018multi,ma2019accurate} or generate meaningful bird's-eye-view (BEV) representations \cite{philion2020lift,reading2021categorical,roddick2018orthographic,srivastava2019learning}, then produce 3D object detection results. 
    These methods have made remarkable progress in M3D.
\begin{figure}[t]
\begin{center}
    \includegraphics[width=0.8\linewidth]{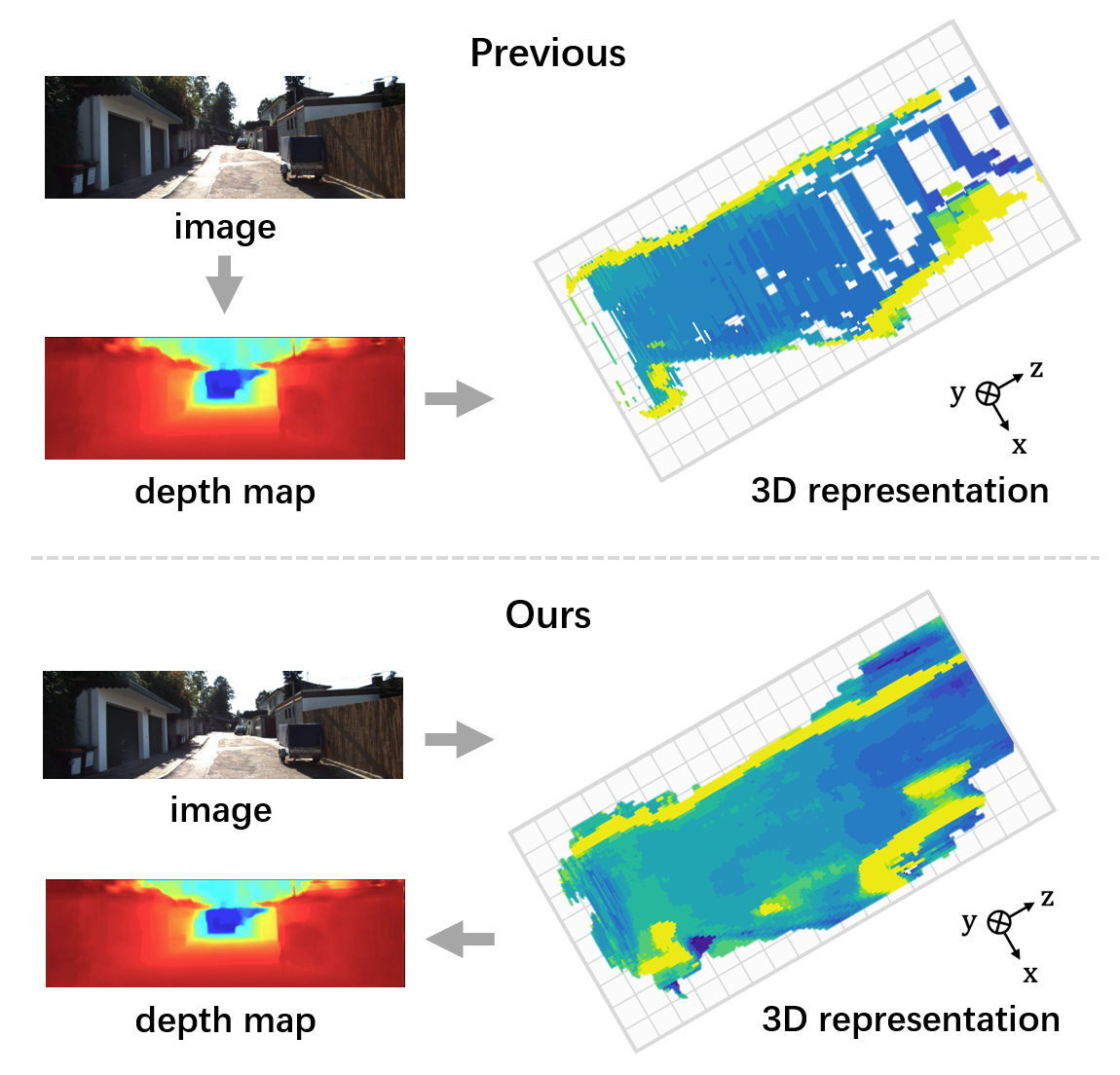}
\end{center}
\caption{
Different intermediate 3D representations.
Previous depth-based methods generate 3D representations by back-projecting estimated depth maps to 3D space, whereas our method predicts implicit 3D representations, and obtains depth estimates through volume rendering.
}
\label{fig:1}
\end{figure}

    Unfortunately, current 3D representations transformed by explicit depths have some limitations.
    First, the lifted features obtained from depth estimates or pseudo-LiDAR exhibit an uneven distribution throughout the 3D space. Specifically, they have high density in close range, but the density decreases as the distance increases. (See top part of Figure \ref{fig:1}).
    Second, the final detection performance heavily depends on the accuracy of the depth estimation, which remains challenging to improve.
    Thus such representations cannot produce dense reasonable 3D features for M3D.
    
    Recent researches in Neural Radiance Fields (NeRF) \cite{barron2021mip,mildenhall2021nerf,zhang2020nerf++} have shown the capability to reconstruct detailed and dense 3D scene geometry and occupancy information from posed images (images with known camera poses).
    Inspired by NeRF and its follow-ups \cite{han2022single,li2021mine,yariv2021volume}, we reformulate the intermediate 3D representations in M3D to NeRF-like 3D representations, which can produce dense reasonable 3D geometry and occupancy.
    To achieve this goal, we combine extracted 2D backbone features and corresponding normalized frustum 3D coordinates to construct 3D Position-aware Frustum features (Section \ref{sss:image2frustum}).
    These 3D features are used to create signed distance fields and radiance fields (RGB color). 
    The signed distance fields encode the distance to the closest surface at every location as a scalar value.
    This allows us to model scenes implicitly by the zero-level set of the signed distance fields. (Section \ref{ss:sdf}).
    We then adopt volume rendering \cite{max1995optical} technique to generate RGB images and depth maps from the signed distance fields and radiance fields (Section \ref{sss:frustumasnerf}), supervising them by original RGB images and LiDAR points (Section \ref{ss:loss}).
    While the previous depth-based methods generate 3D representations based on predicted depth maps, our method generates depth maps based on 3D representations (Figure \ref{fig:1}).
    It is worth noting that our approach is capable of generating dense 3D occupancy (i.e., volume density) without requiring explicit binary occupancy annotations for individual voxels.
    Experiments on KITTI \cite{geiger2012we} 3D benchmark and Waymo Open Dataset \cite{sun2020scalability} show the superiority of our NeRF-like representations in monocular 3D detection.
    
    Our main contributions are threefold:
    \begin{itemize}
        \item 
        We present a novel detection framework, named \textbf{MonoNeRD}, that connects Neural Radiance Fields and monocular 3D detection.
        It leverages NeRF-like continuous 3D representations to enable accurate 3D perception and understanding from a single image.
        \item 
        We propose to use volume rendering to directly optimize 3D representations in detection tasks. 
        To the best of our knowledge, our work is the first to introduce volume rendering for 3D detection tasks.
        \item 
        Extensive experiments on KITTI \cite{geiger2012we} 3D detection benchmark and Waymo Open Dataset \cite{sun2020scalability} demonstrate the effectiveness of our method, which is competitive with previous state-of-the-art works.
        This research presents the potential of 3D implicit reconstruction for image-based 3D perception.
    \end{itemize}
\section{Related Work}
    \subsection{Monocular 3D Object Detection}
        Many impressive researches \cite{reading2021categorical,chen2021monorun,peng2022did} been done in monocular 3D object detection. 
        As the essential information of depth dimension is collapsed in the 2D image, 
        it is an ill-posed but necessary problem of finding an approach to lift the 2D information to 3D space.

        Some Monocular 3D Object Detection methods lift 2D to 3D directly by incorporating the geometric relationship between the 2D image plane and 3D space.
        For example, many methods assume rigidity of objects, aligning 2D keypoints with their 3D counterparts \cite{ansari2018earth,barabanau2019monocular,he2019mono3d++,qin2019monogrnet}.
        Some researchers try to build the bridge between 2D and 3D structures by template matching \cite{chabot2017deep, manhardt2019roi}. 
        These methods usually require additional data (\textit{e.g.}, 3D CAD models, object keypoint annotations). 
        
        Other methods generate intermediate 3D representations to solve the lifting problem. 
        Many approaches \cite{you2019pseudo,ma2019accurate,ma2020rethinking} convert the estimated depth map (always predicted by an off-the-shelf model) to pseudo-LiDAR point cloud representations, and then feed them to sophisticated LiDAR-based detectors. 
        CaDDN \cite{reading2021categorical} learns categorical depth distributions over pixels to lift camera images into 3D space, constructing BEV representations.
        Pseudo-Stereo \cite{chen2022pseudo} uses some sophisticated modules (\textit{e.g.}, stereo cost volume) to achieve better depth estimation.
        These methods typically involve estimating depth maps beforehand and subsequently extracting corresponding 3D representations from them.
        Instead, our method recovers depth maps from the 3D representations and supervises them to improve the detection performance.
        
    \subsection{Neural Implicit Representations}
        Representing 3D geometry with implicit functions has gained popularity in the past few years. 
        Learning-based 3D reconstruction methods \cite{chen2019learning,genova2019learning,mescheder2019occupancy,michalkiewicz2019implicit,park2019deepsdf,saito2019pifu} and their scene-level counterparts \cite{chabra2020deep,chibane2020neural,jiang2020local,peng2020convolutional} demonstrate compelling results but require 3D input and supervision. 
        Neural Radiance Fields (NeRFs) \cite{mildenhall2021nerf} introduce a new perspective to learn continuous geometry representations from posed multi-view images only by volume rendering \cite{max1995optical}. 
        Due to the low efficiency of original MLP-based NeRF, their grid-based variants \cite{hedman2021baking,liu2020neural,yu2021plenoctrees,chen2022tensorf} have been explored to accelerate the rendering process or save GPU memory.
        To circumvent the requirement of dense inputs, some works \cite{yu2021pixelnerf,chen2021mvsnerf,chibane2021stereo,niemeyer2022regnerf} explicitly consider sparse input scenarios and show astonishing results for novel view synthesis, but these representations still need to be optimized per scene. 
        Li \etal \cite{li2021mine} propose MINE in which they combine Multiplane Images (MPI) and NeRF for single-image-based novel view synthesis.
        SceneRF~\cite{cao2023scenerf} uses posed image sequences for training, alleviating the reliance on datasets.
        Wimbauer \etal \cite{wimbauer2023behind} construct a density field to preform both depth prediction and novel view synthesis.
        Unlike our approach, the works mentioned primarily focus on the Novel View Synthesis task, requiring multi-view or sequenced images with camera poses as supervision, which cannot generalize to unseen scenes (except \cite{li2021mine,wimbauer2023behind}).
        Instead, we take advantage of volume rendering to optimize 3D feature representations and improve the performance in M3D task.

\section{Preliminaries}
    Here we introduce some essential preliminaries for our idea. We use the same notation as \cite{yariv2021volume} for the concepts of \textbf{SDF} and \textbf{Volume density}.
    Concerning the space limitation, here we provide a brief overview.
    Please refer to \cite{mildenhall2021nerf,yariv2021volume,max1995optical} for more details.
    
    \subsection{Signed Distance Function (SDF)}\label{ss:sdf}
        For a given spatial point, Signed Distance Function (SDF) is a continuous function that outputs the point's distance to the closest surface, whose sign indicates whether the point is inside (negative) or outside (positive) of the surface.

        Let $\mathbb{R} ^ 3$ represent a 3D space containing some objects, $\Omega \subset \mathbb{R} ^ 3$ refers to the occupied part, and $\mathcal{M} = \partial \Omega$ be the boundary surface.
        Denote the $\Omega$ indicator function by $\mathbf{1}_\Omega$, and the Signed Distance Function (SDF) by $d_\Omega$,
        \begin{equation}
            \mathbf{1}_\Omega(\bm{x}) = 
                \begin{cases} 
                    1 & \text{if } \bm{x}\in \Omega \\ 
                    0 & \text{if } \bm{x}\notin \Omega 
                \end{cases}
        \end{equation}
        \begin{equation}
            d_\Omega(\bm{x}) = (-1)^{\mathbf{1}_\Omega(\bm{x})}\min_{\bm{y}\in\mathcal{M}}\left\Vert \bm{x}-\bm{y}\right\Vert
        \end{equation}
        where $\left\Vert{\cdot}\right\Vert$ is the standard Euclidean 2-norm.
        The underlying surface $\mathcal{M}$ can be implicitly represented by the zero-level set of $d_\Omega({\cdot}) = 0$.
        
    \subsection{Transform SDF to Density}\label{ss:sdf2density}
\begin{figure}[ht]
\begin{center}
\includegraphics[width=\linewidth]{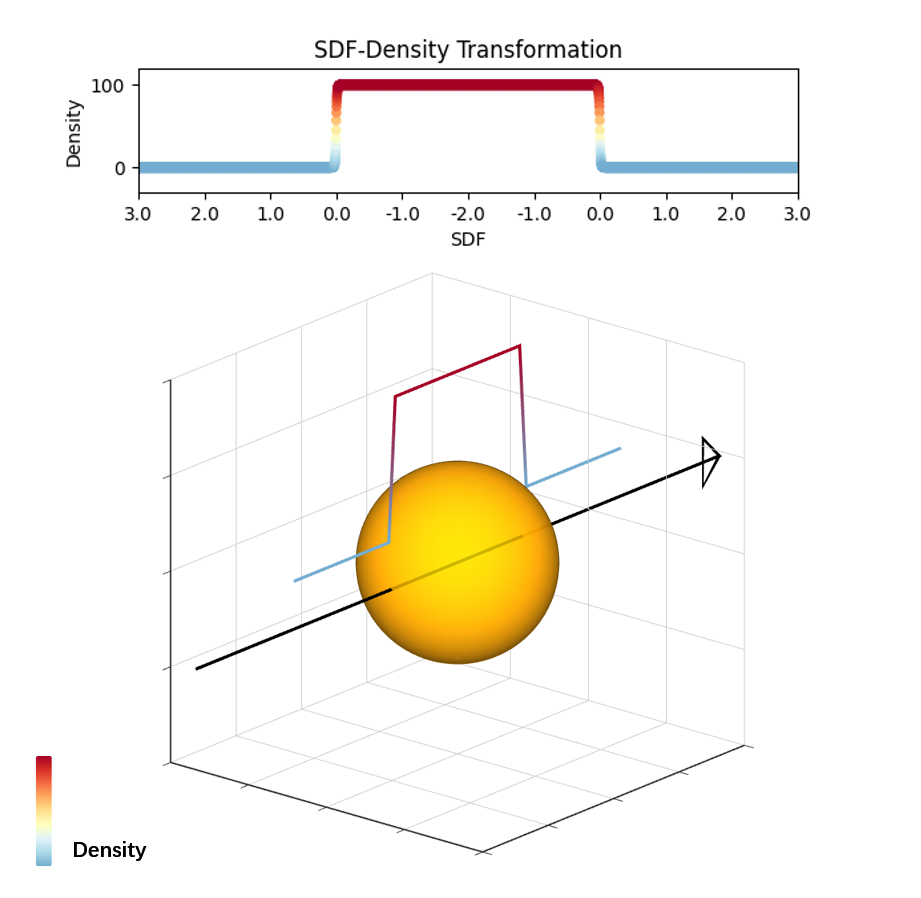}
\end{center}
\caption{Example of transformation from SDF to Density. The black arrow line represents a ray passing through a spherical object; the gradient color line describes how density varies along with SDF values by Equation \ref{eq:sdf2density}, where we set $\beta=0.01, \alpha=\beta^{-1}=100$.}
\label{fig:2}
\end{figure}

        The volume density $\sigma : \mathbb{R} ^ 3\rightarrow \mathbb{R}_+$ is a scalar volumetric function, where $\sigma(\bm{x})$ can be interpreted as the probability that light is occluded at point $\bm{x}$. 
        Previous works \etal \cite{yariv2021volume, yu2022monosdf} suggest modeling the density $\sigma$ using a certain transformation of a learnable Signed Distance Function (SDF), as follows:
        \begin{equation}\label{eq:sdf2density}
            \sigma(\bm{x}) = \alpha \Psi_\beta ({-d_\Omega(\bm{x})})
        \end{equation}
        where $\alpha, \beta > 0$ are learnable parameters and $\Psi_\beta$ is the cumulative distribution function of the Laplace distribution with zero mean and $\beta$ scale,
        \begin{equation}\label{eq:cdf-laplace}
            \Psi_\beta(s) = 
                \begin{cases} 
                    \frac{1}{2} \exp ({\frac{s}{\beta}}) & \text{if } s\leq 0 \\
                    1-\frac{1}{2}\exp ({-\frac{s}{\beta}}) & \text{if } s>0
                \end{cases}
        \end{equation}
        As $\beta$ approaches zero, the density $\sigma$ converges to scaled indicator function of $\Omega$, means $\sigma \rightarrow \alpha\mathbf{1}_\Omega$ for all points $\bm{x} \in \Omega \setminus \mathcal{M}$. The benefits of modeling density in this way have been discussed in \cite{yariv2021volume}.
        We simply set $\alpha = \beta^{-1}$ in our implementations. Figure \ref{fig:2} depicts an example of how density varies along SDF when a ray passes through an object.
    
    \subsection{Volume Rendering}\label{ss:volume-rendering}
        Volume intensity (RGB color) $\bm{c} : \mathbb{R} ^ 3 \rightarrow \mathbb{R} ^ 3$ is a vector volumetric function of spatial position $\bm{x}$, where $\bm{c}(x)$ represents the intensity or color at a certain point $\bm{x}$.
        Given color $c$ and density $\sigma$, the expected color $\bm{C}(\bm{r})$ of ray $\bm{r}(t) = \bm{o} + t\bm{d}$ (where $\bm{o}$ and $\bm{d}$ are the origin and direction of the ray respectively) with near and far bounds $t_n$ and $t_f$ is:
        \begin{equation}
        \begin{split}
            &\bm{C}(\bm{r}) = \int_{t_n}^{t_f}\bm{T}(t)\sigma(\bm{r}(t))\bm{c}(\bm{r}(t))dt, \\
             &\text{where }\bm{T}(t) = \exp\left(- \int_{t_n}^{t}\sigma(\bm{r}(s))ds\right)
        \end{split}
        \end{equation}
        where $\sigma$ refers to the volume density. 
        The function $\bm{T}(t)$ denotes the accumulated transmittance along the ray from $t_n$ to $t$, namely, the probability that the ray (or light) travels from $t_n$ to $t$ without being occluded by any particle.
        In Section \ref{sss:frustumasnerf}, we describe the details to calculate volume rendering numerical integration in our setup.
\begin{figure*}
\begin{center}
\includegraphics[width=\linewidth]{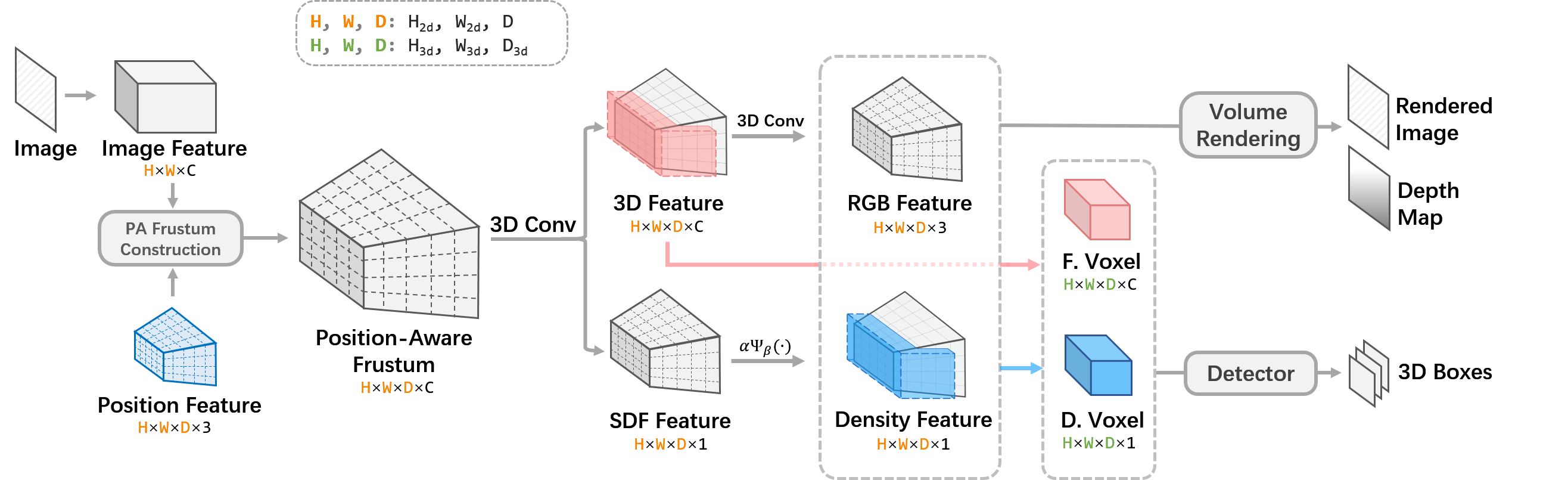}
\end{center}
\caption{Overview of MonoNeRD.
``F. Voxel" and ``D. Voxel" in the figure refer to Feature Voxel and Density Voxel, respectively.
Position-Aware Frustum Features $F_{P}$ are generated from an image $I$ and Positional Frustum Features $F_{pos}$ in a query-based manner (Section \ref{sss:image2frustum}), which are then transformed to 3D features $F^{\prime\prime}$ and signed distance fields $F_{sdf}$. 
After that, the radiance fields $F_{rgb}$ are produced by $F^{\prime\prime}$ (Section \ref{sss:frustumasnerf}). $F_{sdf}$ can be transformed to frustum volume density $F_{density}$ (Equation \ref{eq:sdf2density}) by a scaled cumulative distribution function of Laplace distribution (Equation \ref{eq:cdf-laplace}). 
Volume rendering from $F_{rgb}$ and $F_{density}$ provides supervision for such NeRF-like representations which generate voxel features $V_{3d}$ for M3D (Section \ref{sss:frustum2voxel}).
}
\label{fig:3}
\end{figure*}
\section{Method}\label{s:method}
    \noindent {\textbf{Overview.}}
    The overall framework is illustrated in Figure \ref{fig:3}.
    We follow the network design in LIGA-Stereo \cite{guo2021liga}.
    Note that the stereo part of LIGA-Stereo \cite{guo2021liga} is not applicable for monocular 3D detection (M3D) task.
    We replace the intermediate 3D feature representations with the proposed NeRF-like representations while keeping the 2D image backbone and downstream detection module unchanged.

    Given an input RGB image $I \in \mathbb{R}^{H \times W \times 3}$, it is passed to a 2D image backbone for extracting image features.
    Then we use the 2D image features to generate the proposed NeRF-like 3D representations.
    Specifically, we first build the \textbf{Position-aware Frustum} (See Section \ref{sss:image2frustum}), which is transformed to 3D frustum features and SDF frustum features.
    Such features can be used for obtaining RGB and density features to achieve rendered RGB images and depth maps via \textbf{Volume Rendering} (See Section \ref{sss:frustumasnerf}).
    They are supervised by RGB loss and depth loss to guide previous features, serving as auxiliary tasks in training.
    Furthermore, we use grid-sampling on 3D frustum features and density frustum features to build regular 3D voxel features and corresponding density.
    They are used for forming 3D \textbf{Voxel Features} (See Section \ref{sss:frustum2voxel}), which are fed into the detection modules.
    We elaborate on each part of our method in the following.

    \subsection{NeRF-like Representation}\label{ss:nerf-rep}
        We need to clarify the difference between NeRF and the proposed NeRF-like representation. 
        NeRF encodes a scene with a MLP and has to be trained per scene.
        The proposed method predicts continuous 3D geometry information from the single input RGB image.
        Similar ideas have been explored in researches about single-image-based novel view synthesis. Please refer to \cite{han2022single, li2021mine} for details.
        \subsubsection{Position-aware Frustum Construction}\label{sss:image2frustum}
            To lift 2D backbone features to 3D space without depth, a naive manner is to simply repeat 2D features on different depth planes.
            However, this way does not contain any position information, causing ambiguous spatial features.
            To resolve this problem, we propose to generate position-aware frustum features.
            
\begin{figure}[ht]
\begin{center}
\includegraphics[width=\linewidth]{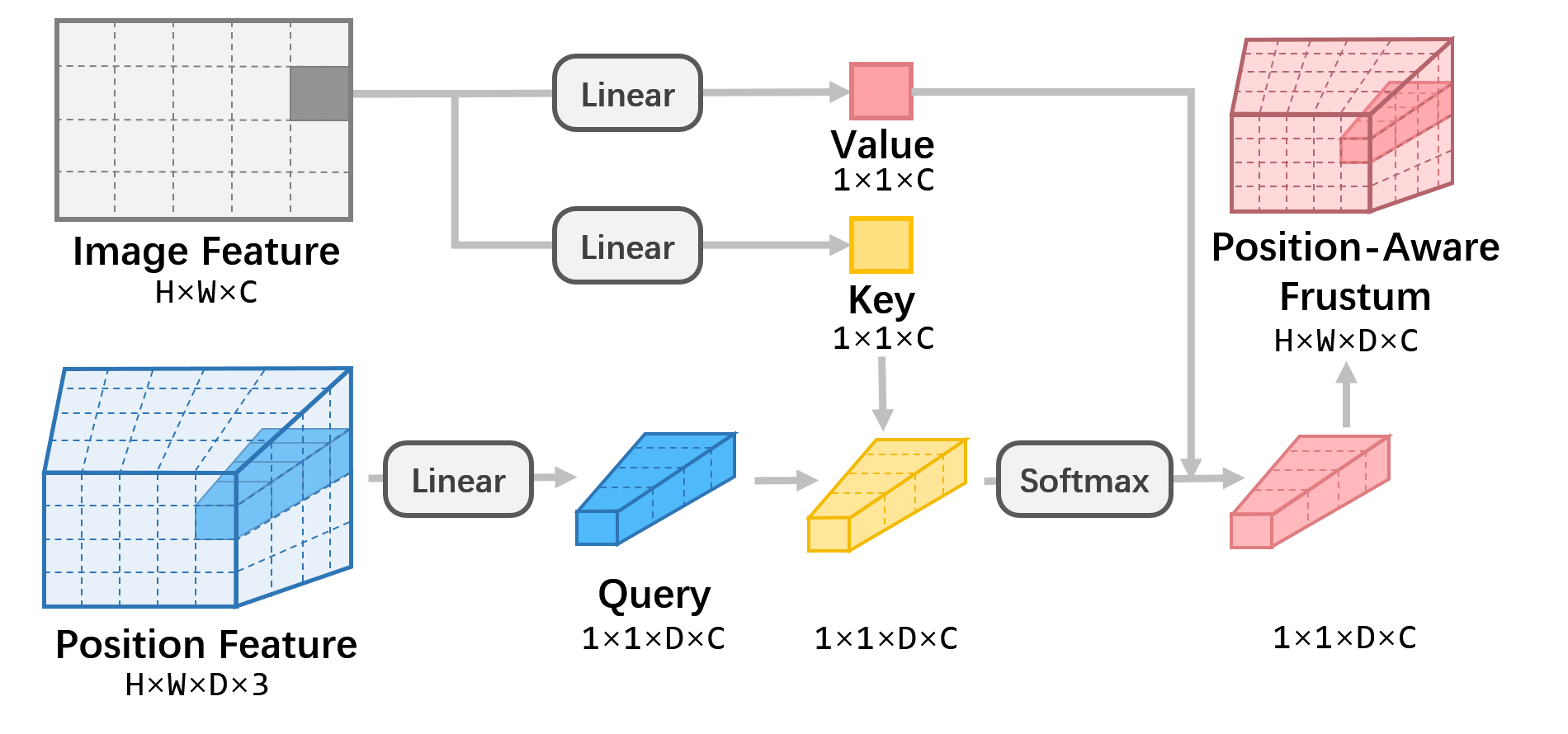}
\end{center}
\caption{The Query-Key-Value mapping procedure for constructing Position-Aware Frustum features.}
\label{fig:4}
\end{figure}

            Taking an RGB image as input, the 2D image backbone produces image features $F_{image} \in \mathbb{R}^{H_{2d} \times W_{2d} \times C}$, where $H_{2d} \text{ and }W_{2d}$ are the height and width of the 2D image feature map, and $C$ is the number of feature channels.
            Such 2D image features are mapped to a camera frustum along with corresponding normalized frustum 3D coordinates in a query-based manner.
            More specifically, given pre-defined near depth $z_{n}$ and far depth $z_{f}$, we sample $D$ planes from this depth range under equal depth intervals with random perturbation as described in \cite{mildenhall2021nerf,li2021mine}.
            Each depth plane consists of frustum 3D position coordinates $p: [u, v, z]^T$ of every pixel point $[u, v]^T \in \mathbb{R}^2$. These coordinates are normalized later. 
            Thus we have 3D positional frustum  features $F_{pos} \in \mathbb{R}^{H_{2d}\times W_{2d}\times D \times 3}$, where $D$ refers to the number of depth planes and $3$ denotes the normalized coordinates.
            To combine 3D position and original 2D image features, we use three linear layers $f_q, f_k, f_v$ to do the Query-Key-Value mapping and attention calculation (in Figure \ref{fig:4}). 
            \begin{equation}
                \begin{split}
                &f_q: F_{pos} \rightarrow Q \in \mathbb{R}^{H_{2d} \times W_{2d} \times D \times C} \\
                &f_k: F_{image} \rightarrow K \in \mathbb{R}^{H_{2d} \times W_{2d} \times C} \\
                &f_v: F_{image} \rightarrow V \in \mathbb{R}^{H_{2d} \times W_{2d} \times C} 
                \end{split}
            \end{equation}
            Therefore, the Position-Aware Frustum features can be calculated by a softmax function along the depth dimension.
            \begin{small}
            \begin{equation}
                F_{P} = \text{Softmax}(QK\text{, dim=D})V
            \end{equation}
            \end{small}
        \subsubsection{Volume Rendering within Frustum}\label{sss:frustumasnerf}
            After constructing position-aware frustum features, we try to encode such features in 3D space to obtain more informative 3D features for volume rendering.
            Towards this goal, we employ two 3D convolution blocks $f_1$ and $f_2$.
            $f_1$ consists of a three-layer 3D-convolution with kernel size 3 and softplus activation.
            It takes position-aware frustum features $F_P\in \mathbb{R}^{H_{2d} \times W_{2d} \times D \times C}$ as input, outputting a same shape feature $F^{\prime}$ except the channel number, as follows:
            \begin{equation}
                f_{1}: F_{P} \rightarrow F^\prime \in \mathbb{R}^{H_{2d} \times W_{2d} \times D \times \left(1+C\right)}
            \end{equation}
            \noindent \textbf{i) SDF and density features.}
            The first channel feature in $F^\prime$ is regarded as SDF feature $F_{sdf}\in \mathbb{R}^{H_{2d} \times W_{2d} \times D \times 1}$.
            The signed distance fields $F_{sdf}$ represent the scene geometry of the input image. For any 3D point $[x, y, z]^T$ in frustum space, we can get its signed distance $s(x,y,z) \in \mathbb{R}^1$ by trilinear sampling in $F_{sdf}$. 
            Furthermore, scene volume density $\sigma$ (or density frustum $F_{density} \in \mathbb{R}^{H_{2d} \times W_{2d} \times D \times 1}$) can be transformed by Equation \ref{eq:sdf2density}:
            \begin{equation}
            \begin{split}
                &\sigma(x, y, z) = \alpha \Psi_\beta ({s(x, y, z)})\\
                &F_{density} = \alpha \Psi_\beta(F_{sdf})
            \end{split}
            \end{equation}
            The density will be used for rendering depth maps and weighing the voxel features.
            
            \noindent \textbf{ii) RGB features.} 
            The remaining $C$ channel feature in $F^{\prime}$, called $F^{\prime\prime}$, is then passed to $f_2$, which consists of a one-layer 3D-convolution with kernel size 3 and sigmoid activation.
            \begin{equation}
                f_2: F^{\prime\prime} \rightarrow F_{rgb}\in \mathbb{R}^{H_{2d} \times W_{2d} \times D \times 3}
            \end{equation}
            The resulting feature is used for approximating the scene radiance field $F_{rgb}$. 
            We can also get color intensity $c(x,y,z) \in \mathbb{R}^3$ by trilinear sampling in $F_{rgb}$.
            
            \noindent \textbf{Rendering from original view.} 
            To save GPU memory, we render downsampled RGB image $\hat I_{low}$ and depth map $\hat Z_{low}$. 
            According to known camera calibration and multiple depth value $\{z_i| i = 1, ...,D\}$, a pixel point $[u, v]^T$ in $\hat I$ can be back-projected to several 3D point coordinates $\{[x_i, y_i, z_i]^T|i = 1, ...,D\}$.
            We can calculate the distance between two successive 3D points as follows:
            \begin{equation}
                \delta_{x_i, y_i, z_i} = \left\Vert{[x_{i+1},y_{i+1},z_{i+1}]^T-[x_{i},y_{i},z_{i}]^T}\right\Vert_2
            \end{equation}
            Denote the distance map at depth $z_i$ as $\delta_{z_i} \in \mathbb{R}^{H_{2d}\times W_{2d}\times 1}$,
            we can easily get the density map $\sigma_{z_i} \in \mathbb{R}^{H_{2d}\times W_{2d}\times 1}$ from $F_{density}$ and color map $c_{z_i} \in \mathbb{R}^{H_{2d}\times W_{2d}\times 3}$ from $F_{rgb}$. 
            Then we can render $\hat I$ by:
            \begin{equation}\label{eq:img-render}
            \begin{split}
                &\hat I_{low} = \sum_{i=1}^D T_i(1-\text{exp}(-\sigma_{z_i}\delta_{z_i}))c_{z_i},\\
            \end{split}
            \end{equation}
            where $T_i = \text{exp} \left(-\sum_{j=1}^{i-1}\sigma_{z_j}\delta_{z_j}\right)$ denotes the accumulated transmittance map from plane 1 to plane $i$.
            Similarly, the low resolution depth map $\hat Z_{low}\in \mathbb{R}^{H_{2d} \times W_{2d} \times 1}$ can be rendered by:
            \begin{equation}\label{eq:depth-render}
                \hat Z_{low}(u, v) = \sum_{i=1}^D T_i(1-\text{exp}(-\sigma_{z_i}\delta_{z_i}))z_i
            \end{equation}     
\begin{figure}[b]
\begin{center}
\includegraphics[width=\linewidth]{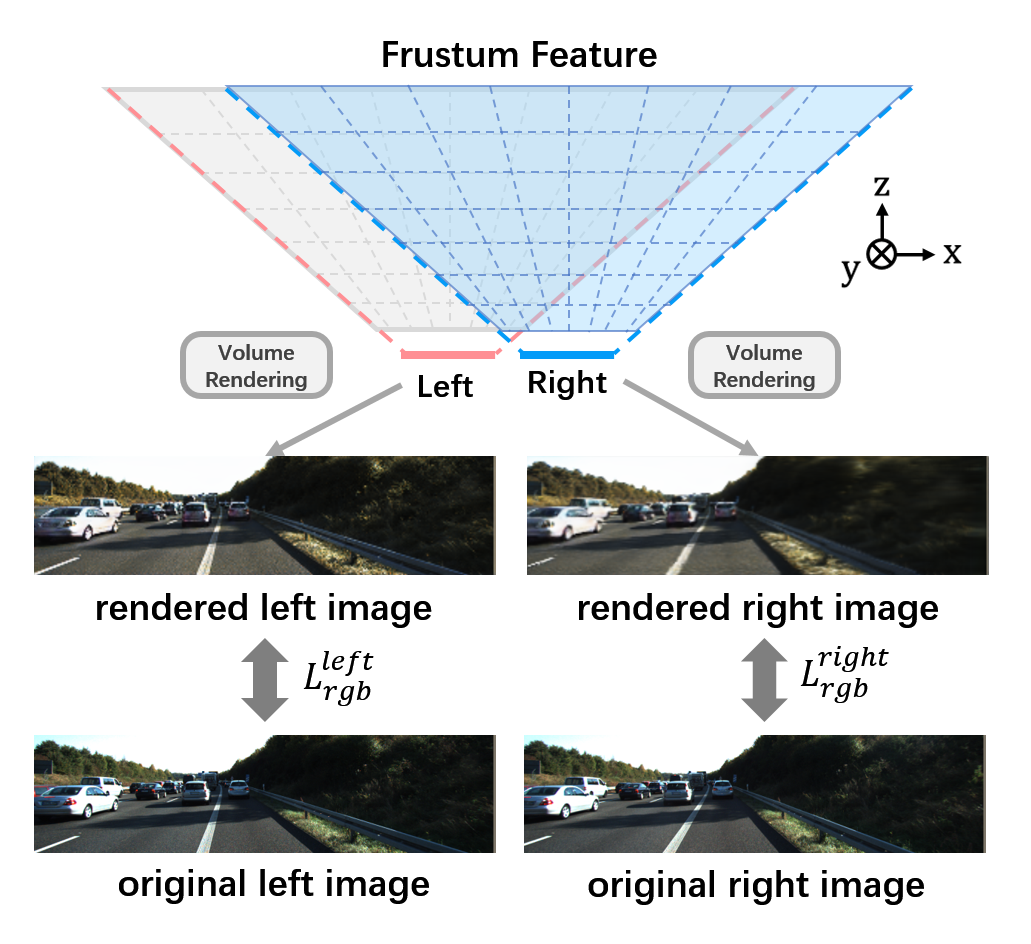}
\end{center}
\caption{
Supervision from two RGB images by using volume rendering. 
We can obtain associated rendered RGB images from frustum density features $F_{density}$ and frustum radiance fields $F_{rgb}$ via Equation \ref{eq:img-render}. 
Under this setting, we do not require explicit depth labels (\textit{i.e.}, LiDAR depths).
Instead, the depth can be implicitly learned by calculating RGB reconstruction loss between original and rendered images. 
}
\label{fig:5}
\end{figure}  

            We simply upsample $\hat I_{low}$ and $\hat Z_{low}$ to original image resolution to get target image $\hat I \in \mathbb{R}^{H\times W\times 3}$ and depth map $\hat Z \in \mathbb{R}^{H\times W\times 1}$.
            Such recovered RGB images and depth maps will be supervised by RGB loss and depth loss (See Section \ref{ss:loss}).
            They are used in training as auxiliary tasks to facilitate the learning process and can be discarded at the inference stage.
     
            \noindent \textbf{Rendering from other views.} 
            Moreover, we can render from other views as long as such views' frustum overlaps with the original view's frustum and their camera calibrations are available. 
            The rendered images can be produced in the same way described above. 
            By calculating the image reconstruction loss from other views' rendered images, our NeRF-like representations can get extra supervisions.
            In Figure \ref{fig:5}, we give an example of taking the stereo images (left and right images) in KITTI \cite{geiger2012we} as volume rendering targets.
            In this way, we can even eliminate the reliance on explicit depth supervisions. 
            Considering the space limitation of the main text, we provided ablation of depth supervision in this supplementary material.

         \subsubsection{Voxel Features Generation}\label{sss:frustum2voxel}
            Aforementioned frustum features cannot be directly used for downstream detection modules as they are irregular in 3D space.
            Therefore, we must transform 3D frustum features to regular 3D voxel features.
            We perform this process of converting frustum to voxel just like CaDDN \cite{reading2021categorical} by leveraging camera calibration and differentiable sampling. 
            First, we define a voxel grid $V \in \mathbb{R}^{H_{3d}\times W_{3d}\times D_{3d}\times 3}$, the sampling points $s_k^v=[x, y, z]_k^T$ are the center of each voxel. 
            We can transform $s_k^v$ to get corresponding frustum sampling points $s_k^f=[u, v, d]_k^T$ according to the camera calibration matrix. 
            Frustum features ($F^{\prime\prime}, F_{density}$) are sampled using sampling points $s_k^f$ with trilinear interpolation to inhabit voxel features ($V^{\prime\prime}, V_{density}$).
            Different from \cite{reading2021categorical}, the spatial resolutions of the frustum grid and voxel grid are not required to be similar because $F_{sdf}$ and $F_{rgb}$ are not limited by spatial resolution.

            Considering that the density indicates 3D occupancy in the 3D space, we use it to enhance sampled 3D voxel features $V^{\prime\prime}$.
            Specifically, we obtain the final 3D voxel features by:
            \begin{equation}
                V_{3d} = V^{\prime\prime} \cdot \tanh(V_{density})
            \end{equation}
            where $\tanh(\cdot)$ is used to scale $V_{density}$ to the range of $[0, 1]$.
            $V_{3d}$ is a NeRF-like representation since it is optimized in the same way as NeRF to learn the implicit scene geometry and occupancy. 
    \subsection{Loss Function}\label{ss:loss}
        Equation \ref{eq:img-render} and \ref{eq:depth-render} build connections between 3D scene representation and 2D observations (\textit{i.e.}, texture and depth). 
        We can therefore optimize the 3D representations with raw 2D RGB images and depth maps. 
        Based on this, there are four terms in the loss function: RGB loss $L_{rgb}$, depth loss $L_{depth}$, SDF loss $L_{sdf}$, and the original loss in the baseline framework LIGA \cite{guo2021liga} $L_\text{LIGA}$.
        The overall loss is formulated as:
        \begin{equation}
            L = \lambda_{rgb} L_{rgb} + \lambda_{depth} L_{depth} + \lambda_{sdf} L_{sdf} + \lambda_{\text{LIGA}} L_\text{LIGA},
        \end{equation}
        where $\lambda_{rgb},\lambda_{depth},\lambda_{sdf},\lambda_{\text{LIGA}}$ are fixed loss weights.
        We empirically set all weights to 1 by default.

        \noindent \textbf{RGB loss.}  The RGB loss consists of two items: Smooth L1 loss $L_\text{smoothL1}$ and SSIM \cite{wang2004image} loss $L_\text{SSIM}$ which are usually used in image reconstruction:
        \begin{equation}
        \begin{split}
            &L_{rgb} = \lambda_{\text{smoothL1}}L_{\text{smoothL1}} + \lambda_{\text{SSIM}}L_\text{SSIM}\\
        \end{split}
        \end{equation}
        
        \noindent \textbf{Depth loss.} Depth map labels are obtained by projecting associated LiDAR points onto the image plane. 
        The depth loss is L1 loss between ground truth sparse depth map $Z$ and predicted depth map $\hat Z$,
        \begin{equation}
            L_{depth} = \frac{1}{N_{depth}}\sum_{u,v}\left\Vert \hat Z - Z \right\Vert_1\\
        \end{equation}
        where $N_{depth}$ denotes the number of valid pixels $(u, v)$ with LiDAR depth.
        
        \noindent \textbf{SDF loss.} The SDF loss encourages $F_{sdf}$ to vanish on geometry surface $\mathcal{M}$,
        \begin{equation}
            L_{sdf} = \frac{1}{N_{gt}}\sum_{x, y, z}(\left\Vert F_{sdf}(x, y, z)\right\Vert_2)
        \end{equation}
        where $N_{gt}$ denotes the number of LiDAR points in the current camera view.

        \noindent \textbf{LIGA loss} We denote the original loss in LIGA \cite{guo2021liga} as $L_\text{LIGA}$. We keep the detection, imitation and auxiliary 2D supervision loss items but remove the stereo depth estimation item to fit the monocular setting. The imitation item is removed in all Waymo \cite{sun2020scalability} experiments for convenience.
        \begin{equation}
            \begin{split}
            &L_\text{LIGA-KITTI} = L_\text{det} + \lambda_{im}L_{im}+\lambda_{2d}L_{2d}\\
            &L_\text{LIGA-Waymo} = L_\text{det} +\lambda_{2d}L_{2d}
            \end{split}
        \end{equation}
\begin{table*}[ht]
\begin{center}
  \small
  \begin{tabular}{l|c|c|ccc|ccc}
    \hline
    \multirow{2}*{Methods} &\multirow{2}*{Reference} &\multirow{2}*{Category} &\multicolumn{3}{c|}{$AP_{BEV}$} & \multicolumn{3}{c}{$AP_{3D}$}\\
    {}  & {} & {} & Easy & Moderate & Hard & Easy & Moderate & Hard \\
    \hline
    D4LCN\cite{ding2020learning}        &CVPR20 &\multirow{2}*{Pretained Depth} &22.51 &16.02 &12.55 &16.65 &11.72 &9.51\\
    DDMP-3D\cite{wang2021depth}         &CVPR21 & {}                            &28.08 &17.89 &13.44 &19.71 &12.78 &9.80\\
    \hline
    Ground-Aware\cite{liu2021ground}    &RAL21  &\multirow{3}*{Directly Regress}&29.81 &17.98 &13.08 &21.65 &13.25 &9.91\\
    MonoRCNN\cite{shi2021geometry}      &ICCV21 & {}                            &25.48 &18.11 &14.10 &18.36 &12.65 &10.03 \\
    MonoEF\cite{zhou2021monocular}      &CVPR21 & {}                            &29.03 &19.70 &17.26 &21.29 &13.87 &11.71 \\
    \hline
    MonoRUn\cite{chen2021monorun}       &CVPR21 &\multirow{4}*{Geometric-based} &27.94 &17.34 &15.24 &19.65 &12.30 &10.58 \\
    GUPNet\cite{lu2021geometry}         &ICCV21 & {}                            &30.29 &21.19 &18.20 &22.26 &15.02 &13.12 \\
    MonoFlex\cite{zhang2021objects}     &CVPR21 & {}                            &28.23 &19.75 &16.89 &19.94 &13.89 &12.07\\
    DCD\cite{li2022densely}             &ECCV22 & {}                            &\textbf{\textcolor{blue}{32.55}} &21.50 &18.25 &\textbf{\textcolor{blue}{23.81}} &15.90 &13.21\\
    \hline
    CaDDN\cite{reading2021categorical}  &CVPR21 &\multirow{4}*{LiDAR Auxiliary} &27.94 &18.91 &17.19 &19.17 &13.41 &11.46 \\
    DD3D\cite{park2021pseudo}           &ICCV21 &{}                             &30.98 &22.56 &\textbf{\textcolor{blue}{20.03}} &23.22 &\textbf{\textcolor{blue}{16.34}} &\textbf{\textcolor{blue}{14.20}} \\
    DID-M3D\cite{peng2022did}           &ECCV22 &{}                             &\textbf{\textcolor{red}{32.95}} &\textbf{\textcolor{blue}{22.76}} &19.83&\textbf{\textcolor{red}{24.40}} &16.29 &13.75 \\
    MonoNeRD (ours)   &--     &{} &31.13  &\textbf{\textcolor{red}{23.46}}  &\textbf{\textcolor{red}{20.97}} &{22.75} &\textbf{\textcolor{red}{17.13}} &\textbf{\textcolor{red}{15.63}} \\
    \hline
  \end{tabular}
\end{center}
\caption{Comparisons for \textit{Car} category on KITTI \textit{test} at IOU threshold 0.7. We obtain the values of other methods from respective papers. We use \textbf{\textcolor{red}{red}} to indicate the highest result and \textbf{\textcolor{blue}{blue}} for the second-highest result. We can see that our method sets a new state of the art.}
\label{table:kitti-test}
\end{table*}

\section{Experiments}
    \subsection{Dataset and Metrics}  
        \noindent
        \textbf{KITTI. }
        The KITTI dataset \cite{geiger2012we} comprises of 7,481  images in the training (\textit{trainval} set) and 7,518 images in the testing (\textit{test} set) with synchronized LiDAR point clouds. 
        In previous works \cite{chen20173d}, the training images were split into a \textit{train} set (3,712 samples) and a \textit{val} set (3,769 samples). 
        We also follow this setting.
        Results on the KITTI dataset \cite{geiger2012we} are measured using IoU-based criteria with a threshold of 0.7 to compute averaged precision over 40 recall values ($AP|_{R40}$) for the car class in both bird's-eye-view (BEV) and 3D tasks.

        \noindent
        \textbf{Waymo.}
        The Waymo Open Dataset \cite{sun2020scalability} consists of 798 training sequences and 202 validation sequences. We follow the setting in CaDDN \cite{reading2021categorical} which only considers the front camera and uses the sampled training set (51,564 samples).
        Results on Waymo \textit{val} are measured by officially evaluation of the mean average precision (mAP) and the mean average precision weighted by heading (mAPH) with IoU criteria of 0.7 and 0.5. The evaluation is performed on two difficulty levels (Level\_1 and Level\_2) and three distance ranges (0 - 30m, 30m - 50m, 50m - $ \infty$). 

    \subsection{Implementation Details}
        \noindent
        \textbf{Training details.}
        Our method is implemented with Pytorch \cite{paszke2019pytorch} framework. We have done all experiments with four NVIDIA 3080Ti (12G) GPUs. The detector is trained using AdamW \cite{loshchilov2017decoupled} optimizer with $\beta_1 = 0.9, \beta_2=0.999$. The batch size is fixed to 4, with 1 sample on each GPU. 
        On KITTI \cite{geiger2012we}, the input size is fixed to $320\times1248$, we train 50 epochs using an initial learning rate of 0.001, and 10 epochs with a reduced learning rate of 0.0001, weight decay is 0.0001. 
        On Waymo \cite{sun2020scalability}, the input size is fixed to $640\times960$, we train 16 epochs using an initial learning rate of 0.001, and 4 epochs with a reduced learning rate of 0.0001, weight decay is 0.0001. 
        
        \noindent
        \textbf{Hyperparameters.}
        On KITTI \cite{geiger2012we}, we sample $D=72$ planes within depth range $[z_n, z_f] = [2, 59.6] (meter)$ in the frustum space. The voxel grid range is $[2, 59.6] \times [-30.4, 30.4] \times [-3, 1] (meter)$ and voxel size is $[0.2, 0.2, 0.2] (meter)$ for the depth, width, and height axis in 3D space, respectively. On Waymo \cite{sun2020scalability} the voxel grid range is $[2, 59.6] \times [-25.6, 25.6] \times [-3, 1] (meter)$ while all other hypeparameters are the same as KITTI experiments.
        
    \subsection{Main Results}
        \noindent
        \textbf{KITTI. }
        Table \ref{table:kitti-test} shows the main results of MonoNeRD on the KITTI \textit{test} set. 
        Our method is competitive with previous methods without any extra data. 
        MonoNeRD achieves the best results under the Moderate and Hard settings in both $AP_{BEV}$ and $AP_{3D}$. 
        Compared to DD3D \cite{park2021pseudo}, which uses a vast private dataset for depth pre-training, we boost the performance from 22.56/16.34 to 23.46/17.13 under moderate setting, and from 20.03/14.20 to 20.97/15.63 under hard setting for $AP_{BEV}/AP_{3D}$, respectively. 
        As for DID-M3D \cite{peng2022did}, we exceed it on both moderate and hard settings with 0.7/1.14 $AP_{BEV}$ and 0.84/1.88 $AP_{3D}$.
        
        \noindent
        \textbf{Waymo.}
        Table \ref{tab:waymo} shows the main results of on the Waymo \textit{val} set. 
        MonoNeRD achieves competitive results without using data augmentation techniques. It has a lightweight backbone (ResNet34) compared to other depth-based methods like CaDDN (ResNet101) and DID-M3D (DLA34). 

        The Results on both KITTI \cite{geiger2012we} hard and Waymo \cite{sun2020scalability} distance ranges (30m - 50m, 50m - $ \infty$) demonstrate the superiority of our method in handling distant and occluded objects.

    \subsection{Ablation Studies}
\begin{table*}[ht]
\begin{center}
    \footnotesize
    \begin{tabular}{l|cccc|cccc}
    \hline
    \multirow{2}*{Methods} & \multicolumn{4}{c|}{3D mAP / mAPH (IoU = 0.7)} & \multicolumn{4}{c}{3D mAP / mAPH (IoU = 0.5)}\\
    {} & Overall &0 - 30m & 30 - 50m & 50m - $\infty$ &Overall &0 - 30m & 30 - 50m & 50m - $\infty$\\
    \hline
    \multicolumn{9}{c}{LEVEL 1}\\
    \hline
    PatchNet \cite{ma2020rethinking}  &0.39 / 0.37   &1.67 / 1.63   &0.13 / 0.12   &0.03 / 0.03   &2.92 / 2.74   &10.03 / 9.75  &1.09 / 0.96   &0.23 / 0.18\\
    CaDDN \cite{reading2021categorical}  &\textbf{\textcolor{blue}{5.03 / 4.99}}   &\textbf{\textcolor{blue}{14.54 / 14.43}}   &\textbf{\textcolor{blue}{1.47 / 1.45}}   &\textbf{\textcolor{blue}{0.10 / 0.10}} &17.54 / 17.31   &\textbf{\textcolor{blue}{45.00 / 44.46}}   &9.24 / 9.11   &0.64 / 0.62\\
    PCT \cite{wang2021progressive} &0.89 / 0.88   &3.18 / 3.15   &0.27 / 0.27   &0.07 / 0.07   &4.20 / 4.15   &14.70 / 14.54   &1.78 / 1.75   &0.39 / 0.39\\
    MonoJSG \cite{lian2022monojsg}  &0.97 / 0.95   &4.65 / 4.59   &0.55 / 0.53   &0.10 / 0.09   &5.65 / 5.47   &20.86 / 20.26   &3.91 / 3.79   &0.97 / 0.92\\
    DEVIANT \cite{kumar2022deviant}  &2.69 / 2.67   &6.95 / 6.90   &0.99 / 0.98   &0.02 / 0.02   &10.98 / 10.89   &26.85 / 26.64   &5.13 / 5.08   &0.18 / 0.18\\
    DID-M3D \cite{peng2022did} & - / -   & - / -   & - / -   & - / -   & \textbf{\textcolor{blue}{20.66 / 20.47}}   &40.92 / 40.60   &\textbf{\textcolor{blue}{15.63 / 15.48}}   &\textbf{\textcolor{blue}{5.35 / 5.24}}\\
    \hline
    MonoNeRD (ours)& \textbf{\textcolor{red}{10.66 / 10.56}} &\textbf{\textcolor{red}{27.84 / 27.57}} & \textbf{\textcolor{red}{5.40 / 5.36}} & \textbf{\textcolor{red}{0.72 / 0.71}} & \textbf{\textcolor{red}{31.18 / 30.70}} &\textbf{\textcolor{red}{61.11 / 60.28}} & \textbf{\textcolor{red}{26.08 / 25.71}} &\textbf{\textcolor{red}{6.60 / 6.47}}\\
    \hline
    \multicolumn{9}{c}{LEVEL 2}\\
    \hline
    PatchNet \cite{ma2020rethinking}  &0.38 / 0.36   &1.67 / 1.63   &0.13 / 0.11   &0.03 / 0.03   &2.42 / 2.28   &10.01 / 9.73  &1.07 / 0.94   &0.22 / 0.16\\
    CaDDN \cite{reading2021categorical}  &\textbf{\textcolor{blue}{4.49 / 4.45}}   &\textbf{\textcolor{blue}{14.50 / 14.38}}   &\textbf{\textcolor{blue}{1.42 / 1.41}}   &\textbf{\textcolor{blue}{0.09 / 0.09}}   &16.51 / 16.28   &\textbf{\textcolor{blue}{44.87 / 44.33}}   &8.99 / 8.86   &0.58 / 0.55\\
    PCT \cite{wang2021progressive} &0.66 / 0.66   &3.18 / 3.15   &0.27 / 0.26   &0.07 / 0.07   &4.03 / 3.99   &14.67 / 14.51   &1.74 / 1.71   &0.36 / 0.35\\
    MonoJSG \cite{lian2022monojsg}  &0.91 / 0.89   &4.64 / 4.65   &0.55 / 0.53   &0.09 / 0.09   &5.34 / 5.17   &20.79 / 20.19   &3.79 / 3.67   &0.85 / 0.82\\
    DEVIANT \cite{kumar2022deviant}  &2.52 / 2.50   &6.93 / 6.87   &0.95 / 0.94   &0.02 / 0.02   &10.29 / 10.20   &26.75 / 26.54   &4.95 / 4.90   &0.16 / 0.16\\
    DID-M3D \cite{peng2022did} & - / -   & - / -   & - / -   & - / -   &\textbf{\textcolor{blue}{19.37 / 19.19}}   &40.77 / 40.46   &\textbf{\textcolor{blue}{15.18 / 15.04}}  &\textbf{\textcolor{blue}{4.69 / 4.59}}\\
    \hline
    MonoNeRD (ours)& \textbf{\textcolor{red}{10.03 / 9.93}} &\textbf{\textcolor{red}{27.75 / 27.48}} & \textbf{\textcolor{red}{5.25 / 5.21}} & \textbf{\textcolor{red}{0.60 / 0.59}} & \textbf{\textcolor{red}{29.29 / 28.84}} &\textbf{\textcolor{red}{60.91 / 60.08}} & \textbf{\textcolor{red}{25.36 / 25.00}} & \textbf{\textcolor{red}{5.77 / 5.66}}\\
    \hline
    \end{tabular}
    \end{center}
    \caption{Results on Waymo \emph{val} set. We obtain the values of other methods from their respective papers. Our method performs the best.}
    \label{tab:waymo}
\end{table*}
        \noindent
        \textbf{Architectural components.}
        We conduct an ablation study on the car category in KITTI \textit{val} to analyze the effectiveness of the proposed NeRF-like representations. 
        In Table \ref{tab:volume-rendering}, Exp. (a) is the baseline which directly takes features sampled from $F_{P}$ as voxel representations $V_{3d}$ for M3D. 
        In Exp. (b), we use more 3D convolution layers to prevent enhancements from complex model structures.
        Exp. (c) shows that supervision only from a single RGB image is useless and even harmful for detection since a purely single image cannot provide depth information.
        We can see that sparse depth map labels from LiDAR can significantly improve the performance of the detector from Exp. (d).
        The effectiveness of $L_{rgb}$ can be revealed by comparing Exp. (d, e) or Exp. (f, g).
        Finally, SDF loss brings noticeable performance gains, especially on $AP_{BEV}$.
\begin{table}[h]
\begin{center}
  \small
  \resizebox{0.48\textwidth}{!}{
  \begin{tabular}{c|cccc|ccc}
    \hline
    \multirow{2}*{Exp.} & \multicolumn{4}{c|}{Setting} &\multicolumn{3}{c}{$AP_{BEV}/AP_{3D}$}\\
    {}  & 3D Conv. & $L_{rgb}$ & $L_{depth}$ & $L_{sdf}$ & Easy & Moderate & Hard\\
    \hline
    (a) &  &  &  &  & 24.96 / 17.01 & 19.27 / 13.38 &16.89 / 11.73\\
    (b) &\checkmark & & & & 25.64 / 17.22 & 19.75 / 13.65 & 18.01 / 11.99\\
    (c) &\checkmark &\checkmark & & & 24.83 / 16.92 & 19.10 / 13.11 &16.74 / 11.73\\
    (d) &\checkmark &  &\checkmark & & 26.91 / 18.72 & 20.87 / 14.54 & 18.57 / 12.60\\
    (e) &\checkmark &\checkmark &\checkmark & &26.46 / 19.44 & 20.16 / 15.03 &17.59 / 12.83\\
    (f) &\checkmark & &\checkmark &\checkmark &27.94 / 20.28 & 21.44 / 15.32 &18.82 / 13.48\\
    (g) &\checkmark &\checkmark &\checkmark &\checkmark &\textbf{29.03 / 20.64} & \textbf{22.03 / 15.44} &\textbf{19.41 / 13.99}\\
    \hline
  \end{tabular}
  }
\end{center}
\caption{Ablation of network structure and volume rendering loss. ``3D Conv.'': 3D convolution blocks as mentioned in Section \ref{sss:frustumasnerf}; ``$L_{\left(\cdot\right)}$'': use the loss for supervision; ``Exp.'': Experiment.
}
\label{tab:volume-rendering}
\end{table}

        \noindent
        \textbf{Positional aware module.} 
        Based on experiment (g) in Table \ref{tab:volume-rendering}, we replace the query-key-value (QKV) based positional aware module with the concatenation of positional features and back-projected image features.
        The results shows a performance degradation in $AP_{BEV}$ from 29.03/22.03/19.41 to 28.39/21.60/18.94, and in $AP_{3D}$ from 20.64/15.44/13.99 to 19.73/15.16/12.94. 
        This experiment demonstrates the effectiveness of the proposed design.
\begin{table}[h]
\begin{center}
  \scriptsize
   {
  \begin{tabular}{c|c|ccc}
    \hline
     \multirow{2}*{Exp.} &\multirow{2}*{Pos.}&\multicolumn{3}{c}{$AP_{BEV}/AP_{3D}$}\\
    {} & {} & Easy & Moderate & Hard\\
    \hline
    1 &QKV & 29.03 / 20.64 & 22.03 / 15.44 & 19.41 / 13.99\\
    2 &Cat. & 28.39 / 19.73 & 21.60 / 15.16 & 18.94 / 12.94\\
    \hline
  \end{tabular}
  }
\end{center}
\caption{Ablation of the position aware frustum. 
``Pos.'': the injection method of positional information. 
``Cat.'': positional feature concatenation.}
\label{tab:qvk-cat}
\end{table}        
    \subsection{Visualization}
\begin{figure}[hb]
\begin{center}
\includegraphics[width=0.9\linewidth, height=87mm]{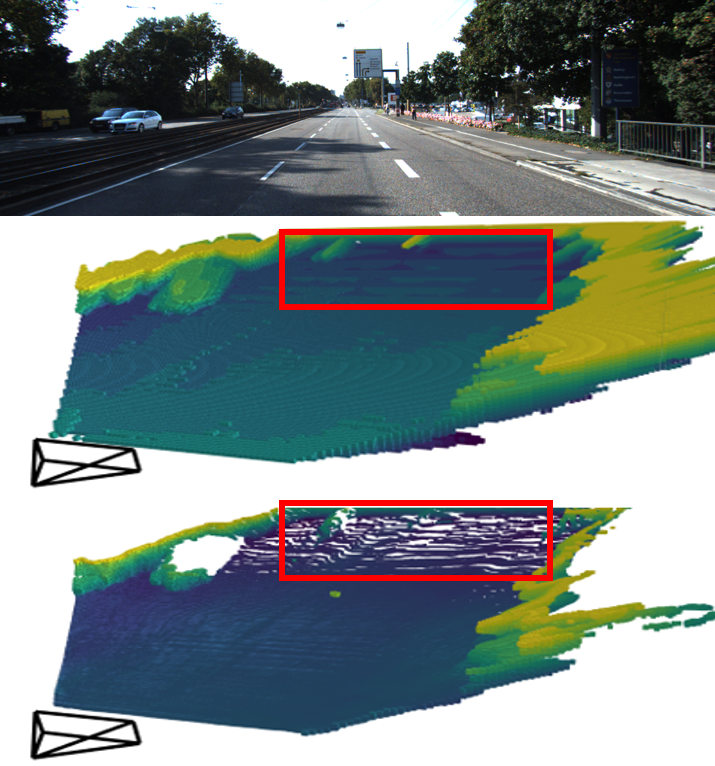}
\end{center}
\caption{
Visualization for depth-map-based representation and the proposed NeRF-like representation. From top to bottom: input image, NeRF-like representation, depth-map-based representation.
Our method generates denser features for distant objects.
}
\label{fig:6}
\end{figure}
        In Figure \ref{fig:6}, we replace the proposed representation with depth-transformed one used in previous works \cite{chen2022pseudo, reading2021categorical} and visualize them for comparison. For depth-transformed category, we first predict the depth map from the input RGB image and apply the same transformation as CaDDN \cite{reading2021categorical} to get its 3D representation. We visualize such representation by back-projecting the estimated depth map to the 3D space. For the NeRF-like representation, we directly visualize the predicted 3D volume density. Our method produces dense and continuous geometry in the distance (as indicated by the red box area). Due to space constraints, we provide additional quantitative and qualitative results in the Supplementary materials. Occupancy results are best viewed as videos, so we urge readers to view our supplementary video.

\section{Conclusion}
    In this paper, we explore how to optimize the intermediate 3D feature representations implicitly for monocular 3D object detection (M3D) without explicit binary occupancy annotations.
    We build a bridge between Neural Radiance Fields (NeRF) and 3D object detection, and propose a novel monocular 3D object detection framework, \textit{i.e.} MonoNeRD,  which produces continuous NeRF-like Representations for M3D. 
    MonoNeRD treats intermediate 3D representations as SDF-based neural radiance fields and optimizes them using volume rendering techniques.
    Extensive experiments show that our method sets a new baseline for monocular 3D detection and has the potential to be extended to other 3D perception tasks.
    
    \noindent \textbf{Limitations and future work.} 
    The performance of our method is highly dependent on the modeling approaches. 
    We are currently limited by the working with bounds modeling (see Section \ref{ss:volume-rendering}), which could fail to predict 3D occupancy when its 2D projected pixel represents sky or other things not within the specified bounds. We believe this is an interesting direction for future research.
\section*{Acknowledgement}
    \noindent This work was supported in part by The National Nature Science Foundation of China (Grant Nos: 62273301, 62273302, 62273303, 62036009, 61936006), in part by the Key R\&D Program of Zhejiang Province, China (2023C01135), and in part by Yongjiang Talent Introduction Programme (Grant No: 2022A-240-G).
    
{\small
\bibliographystyle{ieee_fullname}
\bibliography{egbib}
}

\end{document}